\documentclass[11pt]{article}

\usepackage[a4paper,left=20mm,top=20mm,right=20mm,bottom=25mm]{geometry}

\usepackage{amssymb}
\usepackage{amsthm, amsmath, amsfonts}
\usepackage{mathtools}
\usepackage[mathscr]{euscript}

\usepackage{color}

\usepackage[colorlinks]{hyperref}
\hypersetup{
	citecolor=blue,
	linkcolor=blue,
	urlcolor=blue
}

\usepackage{dsfont} 
\usepackage{graphicx}
\usepackage{grffile}

\DeclareMathOperator\arctanh{arctanh}
\DeclareMathOperator\mean{mean}

\newtheorem{theorem}{Theorem}[section]
\newtheorem{definition}{Definition}[section]
\newtheorem{lemma}{Lemma}[section]

\usepackage{authblk}
\usepackage[font=footnotesize,width=.85\textwidth,labelfont=bf]{caption}

\providecommand{\keywords}[1]{\textbf{\textit{Keywords: }} #1}

\def\be{\begin{eqnarray}}
\def\ee{\end{eqnarray}}
\def\R{\mathbb{R}}
\def\P{\mathbb{P}}
\def\E{\mathbb{E}}

\def\N{\mathcal{N}}
\def\U{\mathcal{U}}

\title{Explainable AI by BAPC - Before and After correction Parameter Comparison}

\author[1]{Florian Sobieczky}
\author[1]{Manuela Gei\ss}

\affil[1]{Software Competence Center Hagenberg GmbH, Hagenberg, Austria}

\date{\ }

\begin{document}
	
\maketitle 

\abstract{
	A local surrogate for an AI-model correcting a simpler 'base' model
	is introduced representing an analytical method to yield
	explanations of AI-predictions. The approach is studied here in the
	context of the base model being linear regression. The AI-model
	approximates the residual error of the linear model and the
	explanations are formulated in terms of the change of the
	interpretable base model's parameters. Criteria are formulated for
	the precise relation between lost accuracy of the surrogate, the
	accuracy of the AI-model, and the surrogate fidelity. It is shown that, assuming a certain maximal amount of noise in the observed data, these criteria induce neighborhoods of the instances to be explained which have an ideal size in terms of maximal accuracy and fidelity. 
}

\bigskip
\noindent

\sloppy

\section{Introduction}
\label{intro}

\subsection{Explainable AI and Regression}

After its initial hype during the past ten years \cite{Dosilovic2018}
\cite{Carvalho2019} \cite{DARPA} \cite{Molnar}, the topic of
Explainable Artificial Intelligence (XAI) has seen strong growth in
its various subfields \cite{Trust}. While various publications are
devoted to fundamental discussions concerning explainability
\cite{John2019} \cite{Ribeiro2016} and metrics to
assess them \cite{metrics}, the maturing of the individual sub-fields lead to a stronger focus on the methodologies necessary to realize
explainability pertinent to the respective sub-fields. 
Recently, research in XAI has started to focus on fostering explainability for learning models which solve regression
tasks \cite{regression}. Similar to the history of statistical
learning theory, in which a full ’theory of the learnable’ \cite{Valiant} was developed \cite{Haussler} only after it had been discussed intensively for classification \cite{Koiran},
regression-specific XAI methods are only developed now, after
explainability of classifying predictive models has seen considerable
attention. While in the case of statistical learning theory it was
necessary to find generalisations of combinatorial terms such as the
VC-dimension \cite{Vapnik} (delivered in the form of the {\em covering number})t , it is necessary for explainability concepts of regression models to generalize what is meant by an interpretation of a decision boundary to the (infinitely) more complex case of an interpretation of a regression function. While, local surrogate models for classifiers may answer the question ’Why did the predictor decide this class to be valid for this given instance?’ they usually cannot naturally be transformed into systems answering questions such as ’Why is the numerical value of the prediction for this instance larger than 0.5?’. The simple idea of turning contour lines of regression functions into decision boundaries only turns the nature of the problem back into a classification problem and ignores the greater wealth of information of the analytical properties of the regression function (such as continuity, steepness, curvature etc.).

In the present paper, we investigate our approach Before and After
prediction Parameter Comparison (BAPC) first proposed for a probit
regression model in \cite{Neugebauer2021} where the power of the method is documented for the case of a physical application under different
degrees of noise. The focus of BAPC is the setting where a pair of
models is combined additively into a total model, where one of them is
interpretable, and the other has black-box character. 
The interpretable one, termed base model, is assumed to approximate the ground truth fairly well (on its own), and to be much more dominant in numerical magnitude than the other model, called ’AI-model’, which gives non-linear corrections to the predictions of the base model. The idea of BAPC to provide explainability is now to mimic the total model locally only by the base model: The parameter change necessary to achieve this for a given instance (and its neighborhood) is the local explanation of the predictive model.

\subsection{Local surrogates at work}

This paper presents a method for making improvements, provided by a non-interpretable AI-model, over given predictions of a model with parameters interpretable in terms of geometric properties of the prediction model as a function of the input data. The approach can be localised in the XAI literature as a model-agnostic, local surrogate of the correcting machine learning model \cite{Molnar}. The problem setup of our approach is often (but not exclusively) found in industrial applications such as predictive maintenance applications \cite{PdM}.  The usefulness of advanced machine learning in the industry of manufacturing processes for mere increase of the predictive accuracy is questionable in the context of applications with many ways of potential improvement in process performance. If there is no hint as to how the improvement of the prediction is conveyed, there is no prescription of a systematic change of the underlying standard procedure, so as to remove the cause for the need of correction \cite{Arrieta}. Furthermore, user acceptance in industry of predictive approaches depends heavily on the ability to interpret machine learning models' solutions \cite{khan}.

Further related work consists in all local surrogate approaches such as LIME \cite{Trust}, which however is designed to explain AI models performing classification tasks. Local surrogates are used heavily in applications for failure control \cite{failureControl}, cancer research \cite{cancer}, epidemiology \cite{epidemiology}, optimal shape design\cite{shapeDesign}, production optimization \cite{productionOptimization}, and borehole research \cite{borehole}. In all these cases the question of locality in the sense of the size of the neighborhood in the input space of a given instance plays an important role.

\subsection{Problem addressed by this paper}

In the light of these observations, we provide a solution for specifically making those non-linear machine learning models interpretable which provide {\em small corrections to regression models}. The interpretations are delivered in the form of characteristic changes of the regression model's parameters. The changes in the parameters are typically also meant to be small, such that the initial choice of the interpretable model is already reasonably accurate and the passing to new, {\em effective} parameters (valid locally) doesn't represent a fundamental change in the picture of the model.

A difficult task in XAI is achieving model-agnostic interpretability on one hand while maintaining model fidelity on the other hand \cite{Ribeiro2016}. It is typically not possible to find good global explanations for a given model, instead local surrogate models are used to provide local explanations. These local surrogate models usually differ in 'how local they are' \cite{Sahra2021}, that is, in the exact definition of the local region for which an explanation is given. In particular, model-agnostic procedures such as LIME or SHAP assume the existence of some proximity measure, with a width parameter ($\sigma$ in \cite{Trust}) resulting in a \emph{Fidelity-Interpretability-Tradeoff}.

Our goal is to provide -apart from the AI model accuracy- a notion of surrogate (i) accuracy and (ii) fidelity, referring to the correctness of the surrogate to the data, on the one hand, and to the AI model, on the other hand. The definitions are supposed to enable the determination of the size of the neighborhood for given instance of interest which is to be explained. We show how a 'weak' form of these definitions leads to a practical method of determining this size for a certain surrogate accuracy to be fulfilled. The result is delivered in a special form of diagram, which we call the \emph{FIDAC-graph}, and allows to control the concurrent existence of surrogate fidelity under given AI model accuracy.

The restrictions that we impose, refer to the model setup: Only model pairs are chosen consisting of a base model together with an AI-model. Only the latter needs interpretation which is delivered as parameter changes of the underlying 'more dominant' base model. In this work, we focus on regression problems. However, the approach may also be extended to classification problems, which will be the topic of forthcoming work. 

The paper is structured as follows:  In Sec.~\ref{sec:BAPC}, we introduce our approach (BAPC) and its general mathematical concept. The workflow of BAPC and a discussion of its local strategy is given in Sec.~\ref{sec:app} by means of the physical example of an  accelerated body due to air-friction. The results are further discussed in Sec.~\ref{sec:disc}.

\section{A new XAI approach: BAPC}
\label{sec:BAPC}

The proposed approach BAPC (``Before and After prediction Parameter Comparison") is explicitly designed for the situation where a given
\emph{interpretable} base model is combined with an error-correcting \emph{non-interpretable} AI-model. This situation is often found in applications, where a process is approximated by a (simple) base model, such as linear regression for instance, and the resulting predictions are then -in a second step- improved by a more complex predictive model. In this context, BAPC aims at explaining the effect of the predictive model in terms of the parameters of the interpretable base model. More precisely, the influence of local corrections performed by a black-box style predictive machine learning model (e.g.\ single layer perceptron) is formulated as the change in the coefficients obtained after fitting the base model a second time -- namely, after the error correction. 

In this section, we establish the theoretical framework of our approach BAPC for regression, discuss crucial aspects such as locality, and define criteria for model accuracy and fidelity. The extension to classification problems will be the topic of future work.

\subsection{Before and After prediction Parameter Comparison (BAPC \cite{Neugebauer2021})}\label{sec:BAPC-workflow}

We consider $\langle X_i, Y_i \rangle$ with $i \in \{0, \dots, n-1\}$, a labeled training set in $\mathcal{X}\times\mathcal{Y}=\mathbb{R}^{p+1}$ with $\mathcal{X}$ a $p$-dimensional  instance space and one-dimensional labels $Y_i\in\mathcal{Y}$. We simplify the notation by setting $\langle \mathbb{X}^n, \mathbb{Y}^n \rangle\coloneqq \{\langle X_i, Y_i \rangle\}_{i \in \{0, \dots, n-1\}}$. This training data will be used to train the models in our approach that can then be used to predict the unknown outcome $Y_n$ of some future input $X_n$.  As shown in Fig.~\ref{fig:schema}, our approach BAPC consists of the three following steps:

\begin{figure}[htb]\vspace*{4pt}
	\centering
	\includegraphics[width=0.78\linewidth]{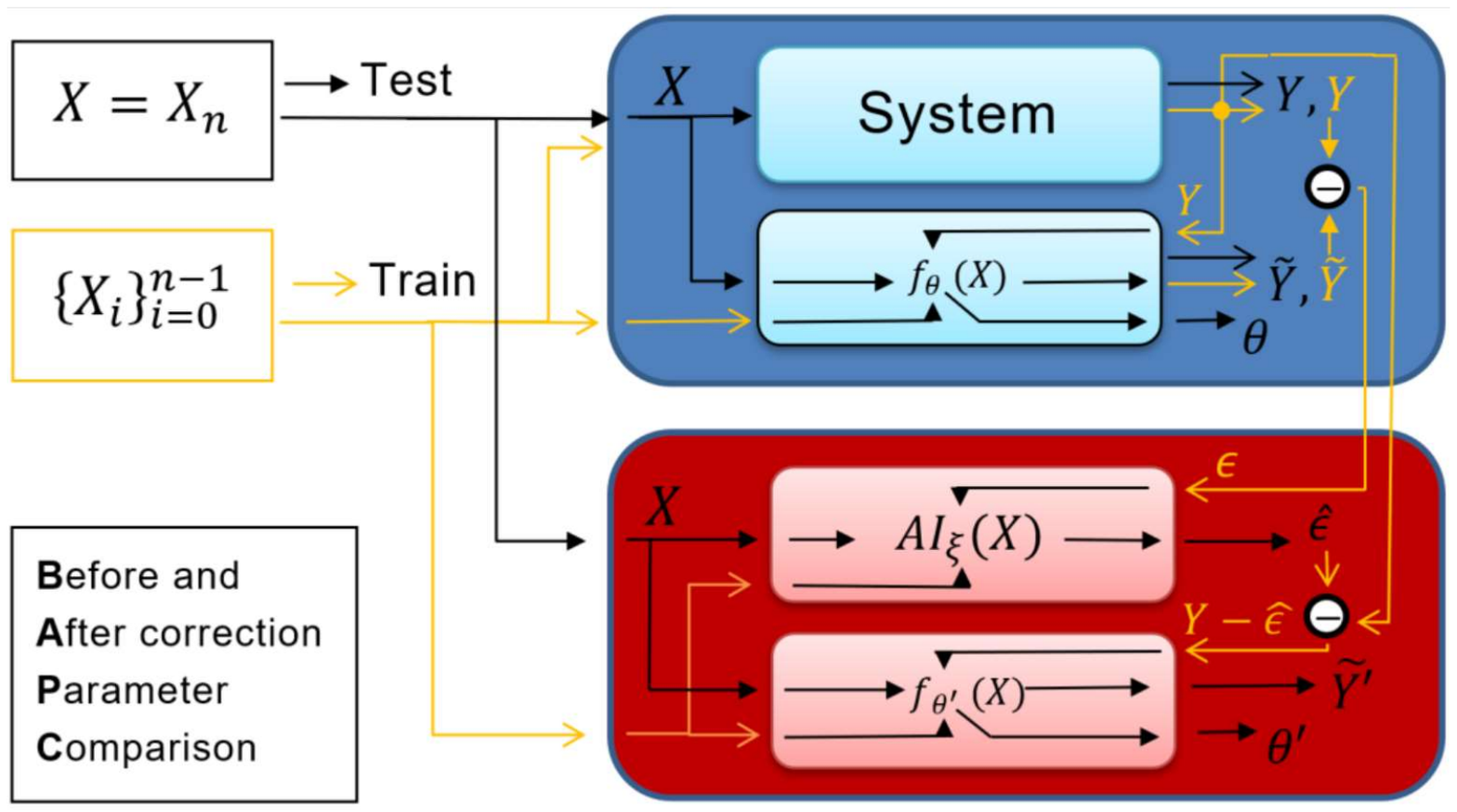}
	\caption{Schematics of BAPC: The 3-step workflow of BAPC is
		shown for both, training on some data set $\langle
		\mathbb{X}^n, \mathbb{Y}^n \rangle$ (yellow lines) and
		prediction for a new input data point $X_n$ (black
		lines). The base model (blue box) estimates the system's
		output $Y$ by $\tilde{Y}$. For training of the base model,
		labeled data pairs $\langle X_i, Y_i\rangle$ are used to
		determine the model parameters $\theta$.  The AI-model
		correction and its corresponding interpretation are obtained
		in the red box: The AI-model is trained on the input values
		$X_i$ together with the residual errors $\varepsilon_i =
		Y_i-\tilde{Y}_i$ as target labels. This determines the
		AI-model's parameters $\xi$ and the corrected 'complete' 
		model $\widehat{Y}_i=f_i\;+\;\widehat{\varepsilon}_i$. In the last step, the base
		model is fitted again, this time using the modified labels
		$\langle X_i, Y'_i\rangle$, where $Y'_i\,=\,Y_i\;-\;\widehat{\varepsilon}_i$ inside a
		predefined neighborhood of $X_n$ and to $Y_i$ everywhere
		else (see also Equ.~(\ref{eq:Yminuseps})). The parameter
		difference $\theta-\theta'$ of the two base model versions
		is used as interpretation of the AI-model's effect. For the
		sake of a simple notation, the indices $i$ of the training
		data are omitted in the graphics. }
	\label{fig:schema}
\end{figure}

\begin{itemize}
	
	\item[(1)] {\bf First application of the base model:} The base model
	is first fitted using the training data $\langle \mathbb{X}^n,
	\mathbb{Y}^n \rangle$ and then applied to $X_n$. The model is given
	as a function $f_\theta:\mathcal{X} \to \mathcal{Y}$ with $\theta$ being a
	vector of parameters having a well-defined geometric meaning:
	\begin{eqnarray}\label{eq:f}
		Y_n\;=\;f_\theta(X_n)\;+\;\varepsilon_n.
	\end{eqnarray}
	The crucial assumption here is that the base model itself is
	interpretable.  The standard examples are linear regression
	(minimization of sum of squared residuals, $\mathcal{Y}=\mathbb{R}$
	and $\theta \in \mathbb{R}^{d+1}$), or probabilistic classification,
	i.e., $\mathcal{Y}=[0, 1]$ and $\theta\in\mathbb{R}^2$ (e.g., probit or
	logistic regression). For the sake of a simple notation, we will write $f$ instead of $f_\theta$.

	\item[(2)]{\bf Application of the AI-correction:} In addition to the
	base model, some non-interpretable \cite{John2019} supervised
	machine learning model $\mathcal{A}_\xi: \mathcal{X}\to
	\mathcal{Y}$ is trained on the labeled data $\xi=\{\langle X_i,
	\varepsilon_i\rangle\}_{i \in \{0, \dots, n-1\}}$, where
	$\varepsilon_i = Y_i - f(X_i)$ is the residual error of the
	base model (therefore $\mathcal{A}_\xi$ is also called the
	difference model).
	
	This yields the corrected prediction $x\mapsto f(x) + \mathcal{A}_\xi(x)$, from which we obtain
	\begin{eqnarray}
		Y_n \;=\; f(X_n) \;+\; \mathcal{A}_\xi(X_n) \;+\;
		\Delta\varepsilon_n.
		\label{eq:step_2}
	\end{eqnarray}
	
	Here, the AI-model
	$\widehat{\varepsilon}_n\coloneqq\mathcal{A}_\xi(X_n)$ estimates the
	residual $\varepsilon_n=Y_n - f(X_n)$ and thereby adds
	additional accuracy to the predictions of the base model
	$f$. In particular, the AI-model is assumed to improve
	the prediction of the base model,
	i.e.\ $|\Delta\varepsilon_n|\le|\varepsilon_n|$ (see Criterion (C1)
	below).

	\item[(3)] {\bf Second application of the base model:} Now, in order
	to render the AI-part of the prediction interpretable in terms of
	the base model, another version of the base model is fitted, but to
	a different set of data, where the AI-correction is taken into
	account in a local neighborhood $\N(X_n)\subseteq \mathcal{X}$ around the instance of interest
	$X_n$.  This leads to the modified training
	data set $\{\langle X_i, Y'_i \rangle\}_{i=0}^{n-1}$, where
	
	\begin{equation}
		Y'_i \coloneqq \begin{cases}
			\displaystyle
			Y_i - \widehat{\varepsilon}_i, & X_i\in \N(X_n)\cap \mathbb{X}^n\\
			Y_i, & X_i\notin \N(X_n)\cap \mathbb{X}^n.		
		\end{cases}
	\end{equation}
	
	Fitting the base model a second time, but this time on $\{\langle X_i, Y_i'\rangle\}_{i=0}^{n-1}$, we obtain
	\begin{eqnarray}\label{eq:Yminuseps} 
		Y'_n \;=\; f_{\theta'}(X_n) \;+\;\varepsilon'_n. 
	\end{eqnarray}
	Again, we simplify the notation by writing $f'$ instead of $f_{\theta'}$. 
\end{itemize}

The region $\N(X_n)$ in Step (3) is typically, assumed to be a star domain with center point $X_n$, that is, for each point $x\in \N(X_n)$ all points on the line segment from $X_n$ to $x$ are contained in $\N(X_n)$. We explicitly emphasize that Equ.~(\ref{eq:step_2}) already gives the
corrected prediction for $X_n$. The third step of BAPC does not aim to further correct this prediction but merely to {\em interpret} the correction
of the AI-model. The surrogate of the AI-model $\mathcal{A}_\xi(X)=\widehat{\varepsilon}$, i.e., the effect of the AI correction, is given by the
difference of the two base models. To understand the correct sign of this correction, note that passing from $Y$ to $Y'$ corresponds to locally removing within $\N(X_n)$ the systematic error of the data which can only be captured by the AI-model. As however the real data contains this error, the corrected prediction of the true observations must contain the step from the corrected to the uncorrected model given by $f(X_n) - f'(X_n)$. In other words,
the difference between $f$ and $f'$ represents this contamination on the level of the base model.  The surrogate model of the AI-prediction $\widehat{\varepsilon}$ is therefore defined as:

\begin{definition}[AI-model Surrogate]
	\label{def:inter}
	Let the corrected prediction $\widehat Y_n\in\mathcal{Y}$ for some $X_n\in\mathcal{X}$ be given by $f(X_n)+\mathcal{A}_\xi(X_n)$ for a pair of supervised learning models $f$ (interpretable base model) and  $\mathcal{A}_\xi$ (residual error predictor). The change in the base model due to the AI-correction is locally captured by the AI-model surrogate,  which  is defined as
	\begin{equation}
		\Delta f(x)\coloneqq f(x)-f'(x)
	\end{equation}
	for $x\in \mathcal{X}$. On the parameter level, the AI-model surrogate is described by the difference of the fitted parameters:
	\begin{equation}
		\Delta \theta\coloneqq \theta-\theta'.
	\end{equation}
	
\end{definition}

The fitted parameters, and thereby
also their difference, can by assumption be interpreted in the
context of the base model. Hence, the difference vector of the fitted
parameters $\theta$ and $\theta'$ provides an interpretation (or
explanation) of the effect of the AI-model $\mathcal{A}_\xi$ in
terms of overestimation or underestimation of the parameters by the
base model. More precisely, a negative entry of $\Delta\theta$ in
some given local region, say $\Delta \theta_i=\theta_i -
\theta'_i<0$, corresponds to a smaller predicted parameter
$\theta_i$ compared to $\theta'_i$, i.e., the parameter is smaller
if the AI corrections are included in the data. From this and the fact
that $\Delta f(X_n)$ represents the action $\widehat{\epsilon}_n$ of the AI-correction of $f(X_n)$, we can
conclude that the AI-model corrects the base model's predictions in
the direction of smaller values of the respective parameter. In
other words, for $X_n$, the base model $f$ overestimates
$\theta_i$. Analogously, we can interpret $\Delta \theta_i >0$ as an
underestimation of $\theta_i$ by the base model.   The
explainability of $\mathcal{A}_\xi$ is therefore provided
\emph{locally} at some data point $x\in\mathcal{X}$ by the
comparison of the base model's action on $x$ with the parameters
$\theta$ and $\theta'$ - before and after the correction is applied
(``\textbf{B}efore and \textbf{A}fter correction \textbf{P}arameter
\textbf{C}omparison" - BAPC).  We explicitly emphasize that
Equ.~(\ref{eq:step_2}) already gives the corrected prediction for
$X_n$. The third step of BAPC does not aim to further correct this
prediction but to interpret the correction of the AI-model.

The idea to 'explain' a {\em small} deviation from a state of a
well-interpreted system by the parameters of that system includes the
notion of the change not to be as dramatic as to render the
description of the changed system with these parameters meaningless. A
change induced by the corection is thought to be {\em small} if the
base model's prediction essentially captures the bulk of the true data
and the correction only adapts the precise numerical value of the base
model parameters. The notion of small deviations is captured in
more detail in Def.~\ref{def:criteria}  below (Condition (C2)) by
comparing the base model's change with the AI-model's predictions.

\subsection{Locality of BAPC}\label{sec:criteria}

A crucial point of BAPC is the locality of the corrections in Step (3). This restriction is needed as otherwise any effect that a correction $\widehat{\varepsilon}_n$ may locally have on $\theta'$, and thus on $f'$, could be re-compensated by other corrections elsewhere in the instance space. A typical scenario are variations in data density in different parts of the instance space. In this case, the base model of Step (1) does insufficiently represent sparse regions of the instance space and AI-corrections over the whole instance space would largely dilute or even completely cover the effect of the corrections within these sparse regions. For this reason, we restrict the action space of the AI-model to some local region $\N(X_n)$ around the instance of interest $X_n$ (see Step (3) of BAPC).
In order to determine a suitable size of this region, we impose two criteria guaranteeing 
the accuracy of the AI-model on one hand and the fidelity of the surrogate $\Delta f = f - f'$ on the other hand in some subregion $\mathcal{U}(X_n)\subseteq \N(X_n)$.

\begin{definition}[Strong BAPC-Criteria]
	\label{def:criteria}
	Let $\langle
	\mathbb{X}^n, \mathbb{Y}^n \rangle\coloneqq\{\langle X_i, Y_i
	\rangle\}_{i \in \{0, \dots, n-1\}}$ with $\langle X_i, Y_i
	\rangle\ \in \mathcal{X}\times\mathcal{Y}=\mathbb{R}^{p+1}$ be the training data set used for fitting some interpretable base model $f$ and $X_n\in\mathcal{X}$ be the 'new instance' for which (i) a prediction
	$\widehat Y_n=f(X_n)+\widehat\varepsilon_n$ and (ii) an interpretation of the AI-correction
	$\widehat{\varepsilon}_n$ has to be made. Let  $\N(X_n)$ be the neighborhood around $X_n$,
	used in Step (3) of BAPC. For $(x,y)\in\mathcal{X}\times\mathcal{Y}$, let
	$\varepsilon(x) = y-f(x)$ be the unknown residual,
	$\widehat{\varepsilon}(x)$ its AI-prediction at $x$, and $\Delta
	\varepsilon(x)$ the corresponding testing error. Moreover, let $\mathcal{U}(X_n)\subseteq \N(X_n)$ and $s\coloneqq \mathbb{E}(|\varepsilon(X)| : X\in \mathcal{U}(X_n))$ be the mean of $|\varepsilon(X_i)|$, $X_i\in \\U(X_n)$. Then, if for each $x\in \U(X_n)$ and for some constants $\eta_1,\eta_2\in(0,1]$, it is true that

	\begin{itemize}
		\item[{\bf (C1)}]\emph{(Accuracy)}{
			\begin{equation}\label{eq:C1}
				|\Delta \varepsilon(x)|\le \eta_1 \cdot s
			\end{equation}  
		}
		\item[{\bf (C2)}]\emph{(Fidelity)}{
			\begin{equation}\label{eq:C2}
				\left|\widehat{\varepsilon}(x) - \Delta f(x)\right| \le \eta_2 \cdot s,
			\end{equation}  
		}
	\end{itemize}
	
	we say that $\U(X_n)$ satisfies \emph{$\eta_1$-accuracy} and \emph{$\eta_2$-fidelity}. 
\end{definition}

The Accuracy Condition (C1) guarantees that the AI-model's correction does not deteriorate the overall prediction for all data points in the chosen neighborhood $\U(X_n)$ of $X_n$. Condition (C2) checks for the change in the base model after applying the AI correction not to be different in comparison to the AI-model's prediction. In other words (C2) ensures the fidelity of the base model's change towards the AI-model. 
In addition, these criteria immediately imply that the model surrogate stays within a certain range of the base model's residual:

\begin{lemma}\label{lem:strong}
	Given the setup from Def.~\ref{def:criteria}, let $\U(X_n)$ satisfy $\eta_1$-accuracy and $\eta_2$-fidelity. Then, any $x\in \U(X_n)$ satisfies
	\begin{equation}
		|\varepsilon(x) -\Delta f(x)| \le s\cdot (\eta_1+\eta_2).
	\end{equation}
\end{lemma}
\begin{proof}
	By definition (see Step (2) of the BAPC procedure in Sec.~\ref{sec:BAPC-workflow}), we have $\varepsilon(x)=\widehat\varepsilon(x)+\Delta\varepsilon(x)$. Together with (C1) and (C2), we thus immediately obtain
	\begin{align}
		|\varepsilon(x)-\Delta f(x)| &= |\widehat\varepsilon(x) -\Delta f(x) +\Delta\varepsilon(x)|\\
		&\le |\widehat\varepsilon(x)-\Delta  f(x)| + |\Delta \varepsilon(x)|\\
		&\le s\cdot (\eta_1+\eta_2).
	\end{align}
\end{proof}

Def.~\ref{def:criteria} requires the criteria (C1) and (C2) to be satisfied for each single data point in the region $\U(X_n)$. However, this assumption turns out to be of limited use in practical applications: On the one hand, typically only a (small) subset of data points in $\U(X_n)\subseteq\mathcal{X}\times\mathcal{Y}$ is known. On the other hand, data in real-world applications is usually noisy. Highly noise-perturbed data points $(X_i, Y_i)$ are not captured well by the AI-model and thus cause fidelity at $X_i$ to
not be fulfilled in the neighborhood even though, typically, the model accuracy	and fidelity are fulfilled for other input vectors close to $X_n$. We therefore additionally propose a weaker form of these criteria:

\begin{definition}[Weak BAPC-Criteria]
	\label{def:criteria-weak}
	
	Consider the probability measure $P$ on the Borel sets of $\mathcal{Z}=\mathcal{X}\times\mathcal{Y}$, the distribution of the independent and identically distributed random variables $\{\langle X_i, Y_i\rangle\}_{i=0}^{n-1}$. Let $\mathbb{P}$ refer to the probability of the product measure $P^n$ on $\mathcal{Z}^n$. Moreover, given the setting and constants of Def.~\ref{def:criteria}, let $\delta_1, \delta_2 \in (0,1)$ some (confidence) parameters. Then, if for each $x\in \U(X_n)$ and with respect to this measure $\mathbb{P}$, it is true that
	
	\begin{itemize}
		\item[{\bf (PC1)}]\emph{(Accuracy)}{
			\begin{equation}
				\mathbb{P}\left[|\Delta \varepsilon(x)|\le \eta_1\cdot s\right] \ge \;1\;\;-\;\; \delta_1, \label{eq:PC1}
			\end{equation}  
		}
		\item[{\bf (PC2)}]\emph{(Fidelity)}{
			\begin{equation}
				\mathbb{P}\left[|\widehat{\varepsilon}(x) - \Delta f(x)| \le \eta_2\cdot s\right] \ge\;1\;\;-\;\; \delta_2,  \label{eq:PC2}
			\end{equation}  
		}
	\end{itemize}
	we say that $\U(X_n)$ satisfies weak
	\emph{($\eta_1,\delta_1$)-accuracy} and
	weak \emph{($\eta_2, \delta_2$)-fidelity}.

\end{definition}

In contrast to the strong
BAPC-Criteria, the weak BAPC-Criteria do not require all but only a
certain proportion of the data points in $\U(X_n)$ to satisfy (C1)
and (C2). Thereby, this definition sets the basis to consider model accuracy and fidelity on the
whole sample space, as opposed to only on the sample itself. While
the underlying distribution is not known explicitly, it is possible to
estimate the probabilities from Equ.~(\ref{eq:PC1}) and
(\ref{eq:PC2}) by the given data set $\langle
\mathbb{X}^n, \mathbb{Y}^n \rangle$ - this makes the definition
useful in an applied setting. In the simulation experiments conducted in the next
section, we will have a closer look at these criteria and discuss
suitable values of $\delta_1$ and $\delta_2$.We will determine the size of the interval $\U(X_n)$ by running the estimates of $\delta_1$ and $\delta_2$ on differently sized test-intervals. The largest interval with acceptable accuracy and fidelity represents the optimal choice of 'neighborhood' on which the same interpretation is valid.

\subsection{Relating surrogate accuracy with AI-model accuracy and fidelity}

The definitions (C1) of AI-model accuracy and (C2) of surrogate fidelity allow making precise statements of the performance of the surrogate. However, it is also useful to control how far off the surrogate model is from the true data. Given a desired surrogate accuracy for some incidence $X_n$, it is of interest to know how to choose a neighborhood of $X_n$ within which this model accuracy holds. As we only have information about the surrogate distribution from the available data samples, estimates of this accuracy will be performed using these samples. In particular, we are interested in the size of an interval placed around a given $X_n \in \R^n$ such that with high probability the surrogate of the correction is smaller than some predefined value:
\be
\P\left[\left|\varepsilon_n\;-\;(f(X_n) - f'(X_n))\right| \le \alpha \right].
\label{eq:surr_err}
\ee

Equivalently, for a desired fidelity, the corresponding quantile
function (inverse function of the one defined in (\ref{eq:surr_err})) yields
the corresponding maximum error of the surrogate. 


The following theorem gives an upper bound for surrogate accuracy. In this theorem and the application that follows, we will
assume (in accordance with the definition of BAPC) that we are
equipped with a labeled data set $\langle X_i, Y_i\rangle\in\mathcal{Z}=\mathcal{X}\times\mathcal{Y}$,
$i\in\{0,\dots, n-1\}$ which allows the definition of the base model
$f(x)$  and the
training of an arbitrary AI-model $\mathcal{A}_\xi: \mathbb{R}^p
\to\mathbb{R}$ with $\widehat{\varepsilon}_i=\mathcal{A}_\xi(X_i)$.
Using this model pair, we consider the point $\langle
X_n,Y_n\rangle\in \mathcal{Z}$, where $X_n$ is chosen
deterministically.  We write $\mathbb{P}[\cdot]$ for the probability
related to the distribution of $Y_n$.

\begin{theorem}[Surrogate Accuracy]\label{thm2.1}
	Given the setup from Def.~\ref{def:criteria-weak}, let $\U(X_n)$ satisfy the weak criteria and $\alpha \ge s \cdot (\eta_1+\eta_2)$. Then, it holds for any $x\in \U(X_n)$:
	\begin{equation}
		\P\left[|\varepsilon(x)-\Delta f(x)|>\alpha\right] \le \delta_1+\delta_2.
		\label{eq:thm2.1}
	\end{equation}
\end{theorem}

\begin{proof} 
	Let $x$ be an instance for which we seek explainability of $\widehat{\varepsilon}(x)$. Note that if $|\Delta \varepsilon(x)|\le s\cdot\eta_1$ and $|\widehat{\varepsilon}(x)-\Delta f(x)|\le s\cdot\eta_2$, then the sum of the terms under the absolute value signs is $|\varepsilon(x)-\Delta f(x)|\le s\cdot(\eta_1+\eta_2)$, by Lemma 2. Conversely, $|\varepsilon(x)-\Delta f(x)| > s\cdot(\eta_1 + \eta_2)$ implies $|\Delta \varepsilon(x)|> s\cdot\eta_1$ or $|\widehat{\varepsilon}(x)-\Delta f(x)|> s\cdot\eta_2$. Therefore,
	\begin{eqnarray}
		\P[|\varepsilon(x)-\Delta f(x)>\alpha] &=& \P[|\Delta\varepsilon(x)\;+\;\widehat{\varepsilon}(x)\,-\,\Delta f(x)|>s\cdot(\eta_1\;+\;\eta_2)]\\
		&\le & \P[|\Delta \varepsilon(x)|>s\cdot\eta_1]\; + \; \P[|\widehat{\varepsilon}(x)\,-\,\Delta f(x)|>s\cdot\eta_2] \\
		&\le& \delta_1\;+\;\delta_2,
	\end{eqnarray}
	where the last inequality sign follows from assumptions (\ref{eq:PC1}) and (\ref{eq:PC2}). 
\end{proof} 

This inequality may seem to be of limited use since $\alpha$ cannot be chosen arbitrarily small. However, as we cannot hope that the surrogate is more accurate than the AI-model, any improvement of its error bound in comparison to the error of the base model restricted to the neighborhood $\U(X_n)$ is valuable and can
be attributed to the prediction of the AI-model.

\begin{figure}[htb]\vspace*{4pt}
	\centering
	\includegraphics[width=0.78\linewidth]{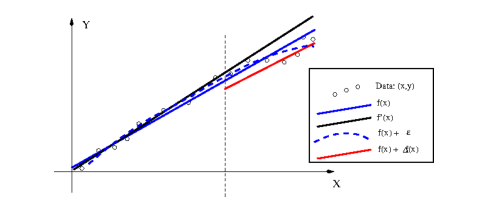}
	\caption{Sketch of BAPC: The linear function depicted in blue
		is the standard linear regression line $f$ for the given
		data (circles). In the region $N$ to the right of the dashed
		vertical line, a departure from an underlying linear Y-vs-X
		relationship is visible, so that an AI-model is applied to
		correct the residual error in this regime. The result of the
		complete (corrected) model is the more accurate prediction
		shown by the dashed blue curve. While the black linear
		function represents the linear regression $f'$ belonging to
		the corrected (linearized) data, the red line is the
		surrogate $f+\Delta f$, locally (in $N$) mimicing the
		action of the complete model in terms of an effective linear
		model. It is seen, that the 'explanation' includes a reduction in the slope of the regression line. The model-fidelity can be verified by the closeness of the linear plus surrogate model (red) to the complete model (dashed blue), while surrogate-accuracy is verified by its closeness to the original data (circles).}
	\label{fig:sketch1}
\end{figure}

As a sketch of BAPC applied to linear regression, consider Figure
\ref{fig:sketch1}, in which a simple regression line (blue) 
approximates data in an unsatisfactory way. The underlying trend of the data is not linear, which a correction yielding an improved approximation
(dashed curve) from a machine learning method which initially cannot
be explained in its effect in the same way as the regression
coefficients yields geometric meaning to the base model approximation
('slope' etc.). 

Within some neighborhood $\N(x_n)$ of the instance $x_n$ taken from a
sample $\{x_i\}$, BAPC now prescribes to  apply the correction by {\bf subtracting} the learned residual error from the data and fit the base model, again (black). This changed linear model $f'$ gives rise to the surrogate $\Delta f(x)=f(x)-f'(x)$
of the AI-correction. Adding it to $f(x)$ shows how much of the AI model can be represented by the surrogate.

The general idea of BAPC is thus one in which the removal of the perturbing anomaly indicates the action of the AI-model, i.e. the understanding of the predictive model $\widehat{\varepsilon}$ comes from learning the difference in $f$ when removing the perturbation. The surrogate must thus be what changes from $f'$ (learned on corrected data) towards $f$ (learned on true data). This explains the sign of $\Delta f$, which is {\em added} to $f$, to explain $f+\widehat{\varepsilon}$.


\section{Application: The departure of linear growth of acceleration due to air-friction}
\label{sec:app}

Using a well-known example from physics, we demonstrate in this section the workflow of BAPC, further discuss our local strategy, and compare data sets with different amount and type of noise.

\subsection{Description of the Experiment}
\label{sec:drag_descr}
The velocity of an object falling through a medium of low density with initial velocity $v_i$  is described by the differential equation
\begin{equation}
	\frac{dv}{dt} = g - \frac{\rho A C_d}{2m} v^2 
\end{equation}
with the solution
\begin{equation}
	v(t)=v_t\cdot \tanh \left(t\cdot \frac{g}{v_t} + \arctanh \left(\frac{v_i}{v_t}\right)\right),
	\label{eq:v}
\end{equation}
where $g$ is the gravitational acceleration and $v_t$ the terminal velocity \cite{Bergman}. The terminal velocity is the maximum velocity that is attained once the acting forces are at equilibrium, and can be calculated by
\begin{equation}
	v_t=\sqrt{\frac{2mg}{\rho A C_d}},
\end{equation}
where $m$ is the object's mass, $\rho$ the fluid's density, $A$ the projected area of the object, and $C_d$ the drag coefficient. 

Using these formulas, we generate a labeled data set $\langle t_i, v_i \rangle$, $0\le i \le 499$, of 500 time points and the corresponding velocity of the falling object by randomly drawing time points $t_i$ from $[0,3]$ and then calculate $v_i$ using Equ.~(\ref{eq:v}). As parameters we choose $g=9.81$, $m=10$, $\rho=1.2$ (corresponding to the air at sea level), $A=1$, $C_d=0.47$ (corresponding to a sphere), and $v_i=0$. 
This procedure generates a data set without noise. For comparison of BAPC with different noise levels, we additionally introduce noise of the form $v_i + \sigma\cdot \mathcal{Z}_i$, where $\mathcal{Z}_i$ is either normally distributed, i.e.\ $\mathcal{Z}_i\sim\mathcal{N}(0,1)$, or uniformly distributed, i.e.\ $\mathcal{Z}_i\sim\mathcal{U}([-1,1])$. Finally, we randomly split the data set into 400 samples $\langle\mathbb{T}_t, \mathbb{V}_t\rangle$ for training and 100 samples $\langle\mathbb{T}_v, \mathbb{V}_v\rangle$ for validation.

The BAPC workflow proposed in Sec.~\ref{sec:BAPC-workflow} is carried out with linear regression as base model and a neural network with one hidden layer of 32 nodes as the AI-model (MLPRegressor from the Scikit-learn library, activation: \emph{relu}, solver: \emph{lbfgs}). For the third step of BAPC, we choose $X_n=2.5$ together with a symmetric neighborhood, more precisely the interval $\N(X_n)=\{X: |X_n-X|\le 0.5\}=[2,3]$.  The results are shown in Fig.~\ref{fig:drag_local} for the validation data set $\langle\mathbb{T}_v, \mathbb{V}_v\rangle$.

Moreover, we check Conditions (C1) and (C2) (see Sec.~\ref{sec:criteria}) for the data sets with normally distributed noise and symmetric intervals $\U(X_n, r)=\{X: |X_n-X|\le r\}=[2.5-r, 2.5+r]$ with different values for $r\in[0, ..., 0.5]$.
We  then estimate $\delta_1$ in (PC1) as well as $\delta_2$ and (PC2) by checking, for different values of the radius $r$, what fraction of points in $\U(X_n)$ satisfies (C1) resp.\ (C2). In this context of weak criteria, we repeat the whole BAPC procedure for 100 randomly generated data sets by using different splits for generating the training and validation data sets from $\langle t_i,v_i\rangle$. The results are summarized in Fig.~\ref{fig:criteria_no_noise} and \ref{fig:criteria_noise} for $\eta_1=\eta_2=1$. We finally analyze to what extent Thm.~\ref{thm2.1} cab be applied to determine an optimal choice for $\eta_1$ and $\eta_2$.

\subsection{Workflow and AI-model interpretation in BAPC}

\begin{figure}[ht!]
	\centering
	\includegraphics[width=1\textwidth]{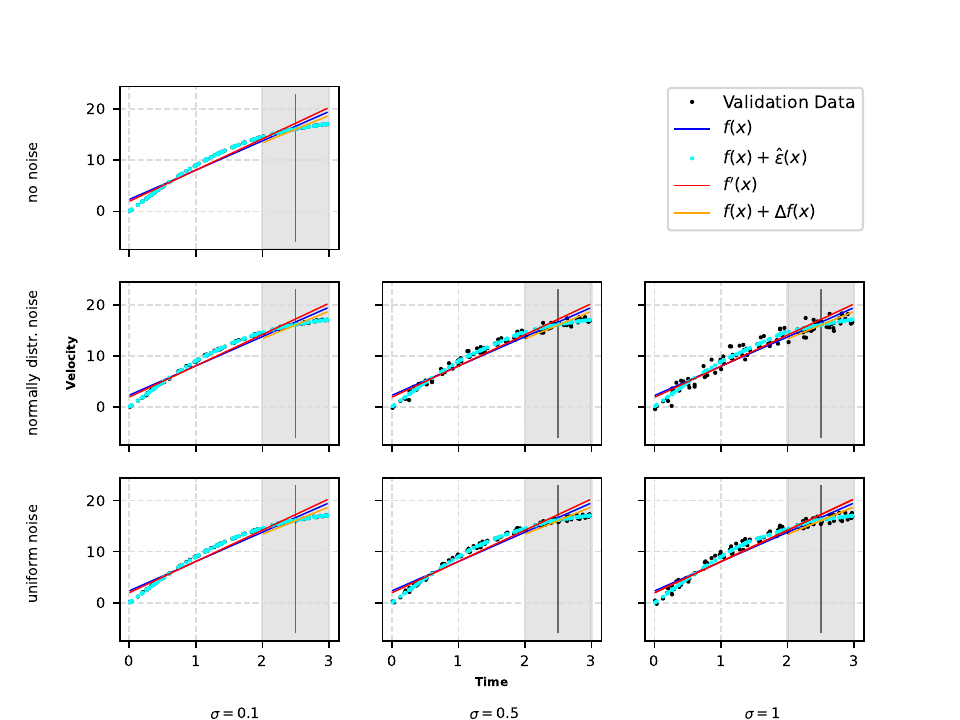}
	\caption{The three steps of BAPC with different noise level. The labeled validation data set (black dots) consists of 100 randomly generated time points $t_i$ in $[0,3]$ and corresponding velocities $v_i$, $0\le i\le 99$, where $v_i$ has been generated by Equ.~(\ref{eq:v}) and is perturbed by different noise levels (no noise, normally distributed noise, and uniformly distributed noise) according to $v_i+\sigma\cdot\mathcal{Z}_i$. The same set of time points has been used for all noise levels. The first fit $f$ of the base model (linear regression) is shown as a (dark) blue line, while (light) blue dots correspond to the base model including the AI-correction $f+\widehat \varepsilon$, where the AI-model is a neural network with one hidden layer containing 32 nodes. The second fit of the base model $f'$ is shown in red, the orange line represents $f+\Delta f$ in $\N(X_n)$. The chosen query data point $X_n=2.5$ and the local neighborhood $\N(X_n)$ are indicated by the gray vertical line and the gray area, respectively. See Sec.~\ref{sec:drag_descr} for further details of this experiment.} 
	\label{fig:drag_local}
\end{figure}

Fig.~\ref{fig:drag_local} shows the models resulting from the three
fitting steps of BAPC for different type of noise and for the
validation data set $\langle\mathbb{T}_v, \mathbb{V}_v\rangle$. Not
surprisingly, due to the concave shape of the falling object's
velocity curve on the interval $\N(X_n)$, the \emph{global} fit $f$ of
the linear regression base model (Step (1)) is hardly a good
approximation of the target data $\mathbb{V}_v$. This is true in
particular for increasing time points (larger than 2), no matter to
what extent the data set has been perturbed.  However, the error of
$f$ is captured well by a well-trained AI-model in Step (2), even for
a highly perturbed data set and particularly for the noiseless data
(the light blue dots practically coincide with the black dots in the
upper left plot in Fig.~\ref{fig:drag_local}). As discussed in
Sec.~\ref{sec:BAPC-workflow} (cf.\ Def.~\ref{def:inter}), the
explanation of the AI-model correction is given by $\Delta
\theta=\theta -\theta'$. In this context we are primarily interested
in the corresponding difference $\Delta \theta_s$ of the slopes, from which we can infer a potential
``over-'' or ``underestimation'' of $\theta_s$. For the interval $\N(X_n)$ (indicated
as gray area in Fig.~\ref{fig:drag_local}), it can be seen that $\Delta
\theta_s <0$, i.e., the acceleration of the falling object is predicted
to be larger after removal of the AI-model's correction. In other
words, inclusion of the AI corrections results in a smaller
acceleration, from which it can be concluded that the AI-model acts to
decrease the acceleration, that is, the acceleration $\theta$
is overestimated by the base model $f$ in the local region $\N(X_n)$. The function $f+\Delta f$ shows that the corrected base model follows the trend of the AI-model in the region of interest $[2,3]$, even though it cannot grasp the entire complexity of the AI-model.

\subsection{Discussion of BAPC-Criteria} 

\begin{figure}
	\centering
	\includegraphics[width=\textwidth]{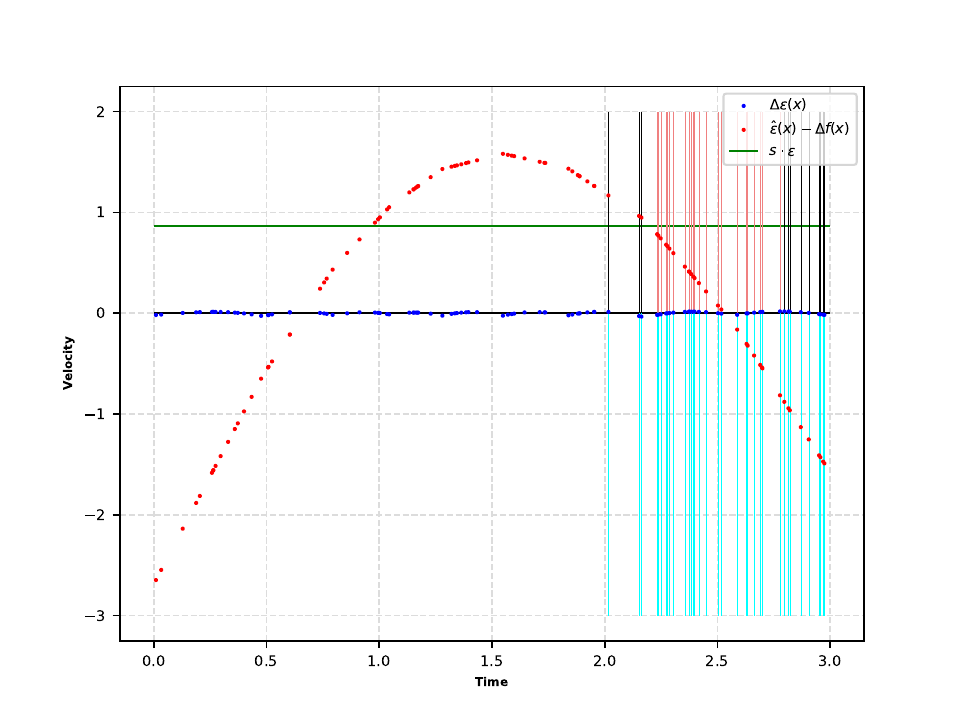}
	\caption{Accuracy and Fidelity (no noise).
		Using the data set without noise, we check the two Criteria (C1) and (C2) for the validation data set in $\U(2.5)=\N(2.5)=[2,3]$. The blue resp.\ red dots indicate the left-hand side of the inequality of (C1), resp.\ (C2) for all validation data points. The right-hand side of the inequalities is represented by the green horizontal line, where we have chosen $\eta_1=\eta_2=1$. In contrast to Def.~\ref{def:criteria}, which uses the absolute values, the real number values - i.e., including the correct sign - are shown. The blue (below x-axis) resp.\ red (above x-axis) vertical lines indicate those data points in $\U(2.5)$, for which Condition (C1) resp.\ (C2) is satisfied. Black lines indicate that these conditions are not satisfied.	
	}
	
	\label{fig:criteria_no_noise}
\end{figure}

\begin{figure}[ht!]
	\centering
	\includegraphics[width=1\linewidth]{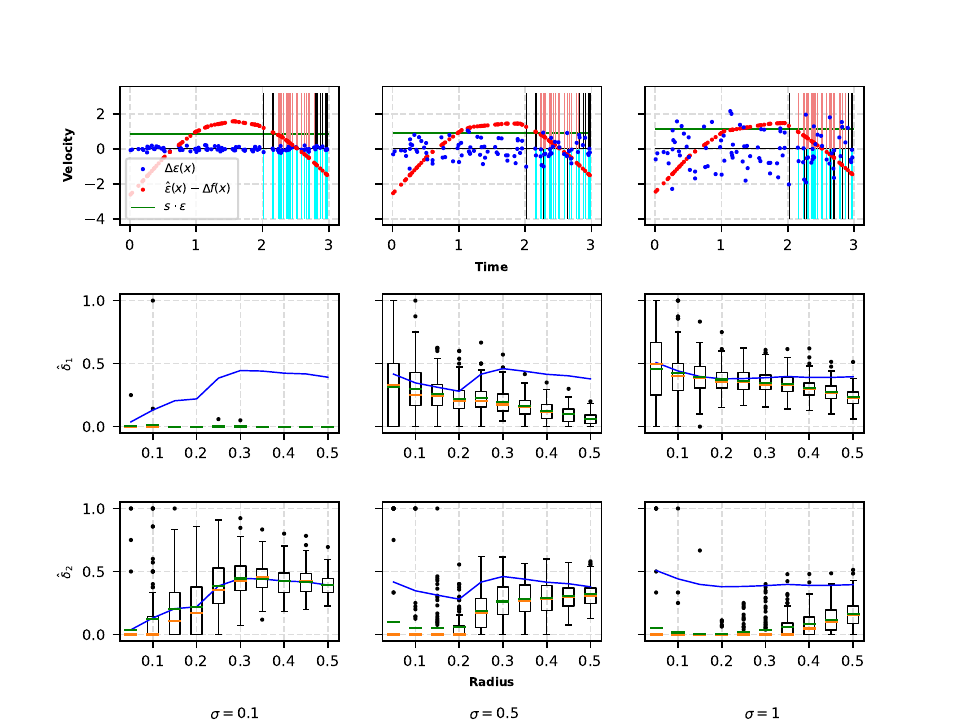}
	\caption{Accuracy and Fidelity (normally distributed noise). \textit{Top row:} 
		Using the same data set (with normally distributed noise and three different noise levels $\sigma=0.1$, $\sigma=0.5$, and $\sigma=1$) and setup  as in Fig.~\ref{fig:drag_local}, we checked  the strong BAPC Criteria (C1) and (C2) for $\U(2.5)=\N(2.5)=[2,3]$. The blue resp.\ red dots indicate the left-hand side of the inequality of (C1), resp.\ (C2) for all validation samples. The right-hand side of the inequalities is represented by the green horizontal line, where we have chosen $\eta_1=\eta_2=1$. In contrast to Def.~\ref{def:criteria}, which uses the absolute values, the real number values - i.e., including the sign - are shown. The blue (below x-axis) resp.\ red (above x-axis) vertical lines indicate those points in $\U(2.5)$, for which Condition (C1) resp.\ (C2) is satisfied. Black lines indicate that these conditions are not satisfied.
		\textit{Middle/bottom row:} Using different values for the radius $r$, the fraction of data points in $U_r(2.5)\coloneqq [2.5-r, 2.5+r]$ \textbf{not} satisfying (C1) (middle row) resp.\ (C2) (bottom row) is shown as boxplots over 100 randomly generated data sets (cf.\ Sec.~\ref{sec:drag_descr}). The mean is shown in green, the median in orange. This fraction can be used as an estimate of $\delta_1$ resp.\ $\delta_2$ in the weak BAPC Criteria (PC1) resp.\ (PC2). The blue line indicates the minimum of the sum $\mean(\hat\delta_1)+\mean(\hat\delta_2)$. Again, the same data set with normally distributed noise and three different noise levels from previous experiments has been used.}
	
	\label{fig:criteria_noise}
\end{figure}

Fig.~\ref{fig:criteria_no_noise} summarizes the results  for the Criteria (C1) and (C2) for the unperturbed data based on $\U(X_n)=[2,3]$, where $f'$ has been generated using $\N(X_n)=[2,3]$.  Note that only the left-hand side of the inequality (C2) is influenced by the choice of  $\N(X_n)$ since, in contrast to $\varepsilon$ and $\Delta \varepsilon$, the refitted base model $f'$ depends on $\N(X_n)$. We observe that the Accuracy Condition (C1) is satisfied for all validation data points in  $[0,3]$ reflecting the fact that the AI correction strongly improves the base model's prediction on a global level. In contrast, the Surrogate Fidelity Condition (C2) is not satisfied for all validation samples in $\U(X_n)$. At the upper end of $\U(X_n)$, where the base model is not a good fit for the samples, the AI correction is large in comparison to the model surrogate $\Delta f$ (cf.\ Fig.~\ref{fig:drag_local}) and thus, (C2) is not satisfied. At the lower end of $\U(X_n)$, the AI correction and $f'$ lie on the same site of $f$ (in this case above), in other words $f'$ does not follow the trend of the corrected predictions from Step (2) of BAPC. Whenever this is too expressed, as it is the case around $x=2$ opposed to e.g.\ $x=2.3$, this causes (C2) to not be satisfied.

\begin{figure}[htb]
	\centering
	\includegraphics[width=\textwidth]{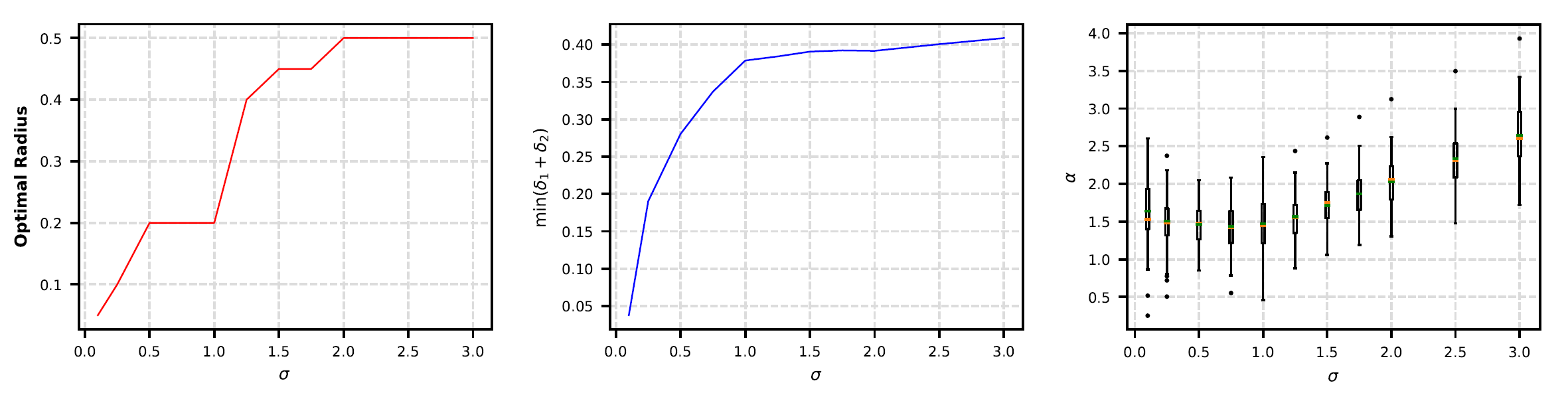}
	\caption{Optimal Neighborhood.  \textit{Left/Middle:} Based on Fig.~\ref{fig:criteria_noise}, the minimum of  $\mean(\hat\delta_1)+\mean(\hat\delta_2)$ and the corresponding radius is shown for different noise levels $\sigma\in[0.1,3.0]$.
		\textit{Right:} Using Thm.~\ref{thm2.1} and $\delta_1+\delta_2$ from the middle plot, values for $\alpha$ and $\eta=\eta_1=\eta_2$ were calculated for different noise levels $\sigma\in[0.1,3.0]$.}
	\label{fig:thm2.1}
\end{figure}

Not surprisingly, the situation is somewhat different in the presence of noise, where outliers cannot be perfectly predicted neither by the base model nor by the AI-model. 
Fig.~\ref{fig:criteria_noise} (top row) summarizes the results for Conditions (C1) and (C2), based on data sets with different amount of normally distributed noise. Here, both Conditions (C1) and (C2) are not satisfied by a considerable amount of data points in $\U(2.5)=[2,3]$. 
We are thus particularly interested in the weak BAPC Criteria (PC1) and (PC2) as formulated in Def.~\ref{def:criteria-weak} allowing a certain fraction of data points in the local region to  not satisfy the strong BAPC Criteria. In this context, we determined the fraction $\hat\delta_1$ resp.\ $\hat\delta_2$ of points in a local neighborhood $\N(2.5)$ around $X_n=2.5$ that do not satisfy (C1) resp.\ (C2) for different values of the radius $r$ (see Fig.~\ref{fig:criteria_noise} (center/ bottom row)). These values can be considered as estimations for the parameters $\delta_1$ and $\delta_2$ in (PC1) and (PC2). 
Unsurprisingly, both $\hat\delta_1$ and $\hat\delta_2$ increase for higher noise levels. Furthermore, for a given noise level, $\hat\delta_1$ decreases for increasing radius, while on the other hand $\hat\delta_2$ increases.  Clearly, the variance of $\hat\delta_1$ resp.\ $\hat\delta_2$ is higher for small $\U(X_n)$ containing less samples.
A similar behavior can be seen for different values of $X_n$ (see Fig.~\ref{fig:bapc_criteria_2.3_eta_1} and \ref{fig:bapc_criteria_2.7_eta_1} in the appendix for $X_n=2.3$ with $\N(X_n)=[1.6,3]$ and $X_n=2.7$ with $\N(X_n)=[2.4,3]$). We conclude that we cannot minimize both $\hat\delta_1$ and $\hat\delta_2$ at the same time, hence we need some criteria to chose the optimal radius, or in other words the optimal $\U(X_n)$. A straightforward, yet suitable choice seems to be the minimization of $\mean(\hat\delta_1)+\mean(\hat\delta_2)$ (blue line in Fig.~\ref{fig:criteria_noise}), which is also in line with Thm.~\ref{thm2.1} (cf.\ Equ.~(\ref{eq:thm2.1})). Here, $\mean(\hat\delta_i)$ refers to the mean of $\hat\delta_i$ over the 100 different validation data sets used. Fig.~\ref{fig:thm2.1} (left/middle) shows the minimum of $\mean(\hat\delta_1)+\mean(\hat\delta_2)$ and the corresponding optimal radius for different noise levels $\sigma$ from $0.1$ to $3.0$. It can be seen that the optimal radius quickly reaches the maximal possible size of 0.5 for very large levels of noise. This suggests that there is some upper bound for noise levels above which a good  balance between (PC1) and (PC2) cannot be guaranteed. On the other hand, however, we observe that the blue line indicating the minimum of $\mean(\hat\delta_1)+\mean(\hat\delta_2)$ is quite flat for higher noise levels - which is shown for $\sigma=1$ in Fig.~\ref{fig:criteria_noise} and was observed in our experiments for higher noise levels as well (not shown here). 

\begin{figure}
	\centering
	\includegraphics[width=\textwidth]{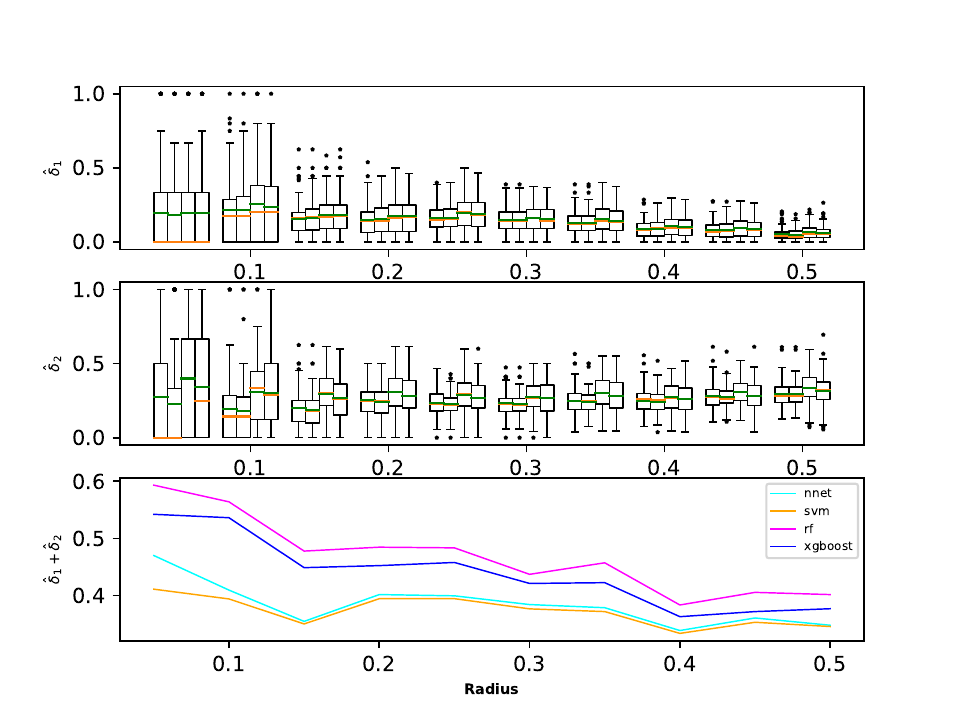}
	\caption{This is the result of the experiment with $\mathcal{U}=\N$: We use normal noise with $\sigma=0.5$ standard deviations, a repetition of the experiment of 100 times, and a cross-validation scheme for each model (data set is the same but the split between training and validation data differs).
		The boxplots from left to right: neural network, SVM, random forest, xgboost. The green line is the mean, the orange the median. It is seen, that the total upper bound (bottom) is best for some value in between 0 and 0.5. Also, all models perform similarily well.}
	\label{fig:diffAI_0.5}
\end{figure}

We now perform an experiment for the case $\N(X_n)=\mathcal{U}(X_n)$ with $X_n=2.5$ and four different AI models: The neural network (nnet)from the previous experiments, a random forest (rf), a support vector machine (SVM), and the gradient boosting algorithm XGBoost. For various sizes of the neighborhood (measured with the radius $r$ for half the interval length centered at $X_n$), a single randomly generated data set of size 100 is used to count the number of points $\langle x_i, y_i\rangle$ for which for each trained predictive model fulfills the accuracy ($\Delta\varepsilon_i$) and fidelity ($\widehat{\varepsilon}_i-\Delta f_i$) condition. It is evaluated with cross-validation with 5 equally sized folds. Fig. \ref{fig:diffAI_0.5} shows the result for a noise level of $\sigma=0.5$: It is seen that
\begin{enumerate}
	\item The results for all four models are very similar (even though SVM and nnet perform best, the bounds for the other two models are not significantly worse).
	\item There is minimum at around $r=0.4$ with an error of about $0.4$ standard deviations of the surrogate approximating the true data on $\N(X_n)$.
	\item Too small values of the radius $r$ make $\widehat{\delta}_1$ too large to reliably estimate the standard deviation $\sigma$.
	\item Large values of the radius $r$ lead to infidelity (high values of $\widehat{\delta}_2$) of the surrogate to the AI-model, of which the non-linear nature becomes more apparent on the larger $\N(X_n)$.
\end{enumerate}

This is seen also very clearly for the run of the experiment with a noise level of $\sigma=0.1$ (see appendix, with a smaller value for the optimal $r$ and estimated error bound $\widehat{\delta}_1\;+\;\widehat{\delta}_2$). Our choice of $\N(X_n)=\mathcal{U}(X_n)$ was made to check the method in the case in which no initial guess for the true range of the 'anomalous' behavior deviating from the linear model's prediction is available. Further experiments have shown that qualitatively similar results are obtained when a fixed (known) size of the influence region $\N(X_n)$ (Step 3 of BAPC) of the perturbation of the data around $X_n$ is chosen, and subsequently different sizes of $\mathcal{U}(X_n)$ are checked in terms of the best explanation $\Delta f(x)$ for $x\in \mathcal{U}(X_n)$. In other words, in this experiment, we are following the principle that choosing a certain size of the set of points where an explanation is reasonably accepted should also be the set of points on which the correction of the data is allowed to take place. Our experimental findings were such that we couldn't reject this principle on the basis of any observed changes in the estimations of $\delta_1$ and $\delta_2$.



As a last step, we analyze if Thm.~\ref{thm2.1} can be used to optimize the choice of $\eta_1$ and $\eta_2$ (reminder: so far, we choose $\eta\coloneqq\eta_1=\eta_2=1$). For this, we use the optimal $\U(X_n)$ and the optimal $\delta_1+\delta_2$ from our previous experiments and apply Equ.~(\ref{eq:thm2.1}) of Thm.~\ref{thm2.1} to calculate $\alpha$ by the $(\delta_1+\delta_2)$-quantile. The resulting $\alpha$'s for the 100 data sets are shown as boxplot in Fig.~\ref{fig:thm2.1} (left). 
Note that for these results, we needed to estimate $s$ also for small values of the radius $r$ where this estimation increases due to there being only few data points left in $\U(X_n)$, there is a natural limitation of the method in the sense that it needs sufficiently large sample sizes $\U(X_n)$. At the same time, a neighborhood of $X_n$ on which the correction $\widehat{\varepsilon}$ changes non-linearly will not be approximated well by (difference of) a linear model. We conclude, that for linear regression there is a certain optimal noise level ($\sigma \sim 0.5$ in our example) for the weak BAPC concept to be working optimally, i.e., to obtain a useful predictive power of the surrogate $\Delta f$ of the true data.



\section{Discussion} \label{sec:disc}

With Definitions 2.2 and 2.3, we are introducing a rigorous definition of explainable AI for a specific class of local surrogates. Namely, those which are applied to predictive models {\em correcting} parametric 'base models'. The base model is meant to act as a ground truth, giving an interpretable, yet usually not sufficiently accurate additive prediction. The complete model is the sum of the (explainable) base model's prediction and the (non-explainable) AI correction.  Explainability of the corrector is then provided for a given instance by the parameter changes necessary to yield the complete model's prediction. We call this vector of parameter changes the explanation vector $\Delta \theta$ which 'tweaks' the base model towards the more accurate prediction of the complete model. 

The local nature of the approach becomes eminent by the
parameter changes being dependent on the specific input data point
(instance $x$). If the assumptions (\ref{eq:PC1}) and (\ref{eq:PC2}) are satisfied in some neighborhood $\N(x)$,
then the same 'explanation' can be used for all instances from this set. The quality of
the base model being less accurate in comparison with the AI-correction
naturally limits its power to replace it. At the same time, it 'explains' the AI-model's action by providing a simplification or 'sketch' of it.

We provide quantitative bounds for the size of $\N(x)$ such that the validity of the vector of parameter shifts in the base model, which yields the complete model's prediction, is guaranteed within the region.

If the quality of model fidelity can be maintained within a region of the instance space which is covered by the union of the neighborhoods $\N(x_i)$ of all instance points $x_i$ of a given sample, then the complete model satisfies an AI-induced increase of accuracy under simultaneous explainability in this region. A second characteristic quality of the complete model is the accuracy of the explanation, termed 'surrogate accuracy', which is also limited by the nature of the base model and the size of the neighborhood on which the explanation vector $\Delta \theta$ is valid.

Altogether, we presented a study of the interplay between the complete model's accuracy, the model fidelity, and the surrogate accuracy on precisely defined neighborhoods $\N(x)$ of an arbitrary instance $x$. A 'complete model prediction' corrects the linear regression estimate of the data and needs to be explained. The definitions PC1 and PC2 provide us with two independent error control parameters ($\delta_1, \delta_2$). They refer to two independent sources of error of the surrogate: The training error $\Delta \varepsilon$ and the lack of fidelity. In Theorem 2.1 they enter into the error bound of the surrogate accuracy as the sum of two independent variables. It shows that typically this bound has a minimum for reasonable sizes defined by the radius $r$ of the neighborhood $\U(x)$ which results from a lack of estimation power of $s$ (for small $r$), and a lack of fidelity (for large values of $r$).

\section{Conclusion and Outlook}

In this work, we have introduced a new model-agnostic approach to provide explainability of an AI-model acting as a corrector of a parametric, interpretable base model. The explainability is established in the framework of local surrogates, more precisely, it is provided in terms of a parameter shift that is \emph{locally} interpretable in the context of the base model. 
We furthermore provided criteria for accuracy of the AI-correction and fidelity of the base model's change, caused by the AI-model, within local neighborhoods. In an application of BAPC to a simulated data set, we investigated the influence of noisy data as well as of the choice of the local neighborhood. Although the AI-model's explanation of BAPC is found to be fairly robust against noise, the level of noise in the data does, not surprisingly, influence both accuracy and fidelity, leading to a distinction between \emph{weak} and \emph{strong BAPC Criteria}. On the other hand, the choice of the local neighborhood does likewise influence accuracy and fidelity and thus also the choice of suitable parameters $\delta_1$ and $\delta_2$ of the weak BAPC Criteria (PC1) and (PC2).
We determined an optimal local region by using a minimization criterion on estimations of $\delta_1$ and $\delta_2$ from multiple experiments. Finally, we provided an upper bound for surrogate inaccuracy that was checked in experiments with the result that under the presence of noise, the upper bound is smallest for some positive $r$ (see Fig.~\ref{fig:thm2.1}). Furthermore, finding the optimal radius (i.e. optimal $\U(X_n)$) for a given noise level, this radius tends to zero for the noise intensity becoming small. However, the probability $\P[|\varepsilon_n\;-\;\Delta f(X_n)|> s(\eta_1+\eta_2)]$ for fixed $\eta_1, \eta_2>0$ and MAE $s=\E[|\varepsilon(X)\,|\,X\in\U(X_n)]$ doesn't converge to zero, but shows to have a minimum at some 'ideal' case-dependent positive noise intensity.

In ongoing work, we are extending BAPC to classification problems and are applying it to classification base models such as decision trees. Once the theoretical framework has been established, BAPC can be applied to a variety of problems, e.g., different predictive maintenance tasks on production data such as 'perturbed' time-series of sensor readings. Furthermore, there is work of BAPC applied to time series \cite{alfredo}. The ability to place high-performing machine learning models (such as neural networks) into the framework of a conventional base model (such as linear or probit regression) is highly attractive as in this case the base model is likely to emerge from the understanding of underlying processes, such as physical or biological processes.
Such a procedure is typical for applications like predictive maintenance, in which "to tune the machine configuration towards less likely faults to happen" (see \cite{Olivan}, Sect.~2) is an Industry 4.0 "methodological criterion".

\section*{Acknowledgements}

This work has been supported by the project 'inAIco' (FFG-Project No. 862019; {\em Bridge Young Scientist}, 2020), as well as the Austrian Ministry for Transport, Innovation and Technology, the Federal Ministry of Science, Research and Economy, and the Province of Upper Austria in the frame of the COMET center SCCH.

\appendix

\section{Detailed fidelity simulation results} \label{app:fidelity_results}

\begin{figure}[h]
	\centering
	\includegraphics[width=\textwidth]{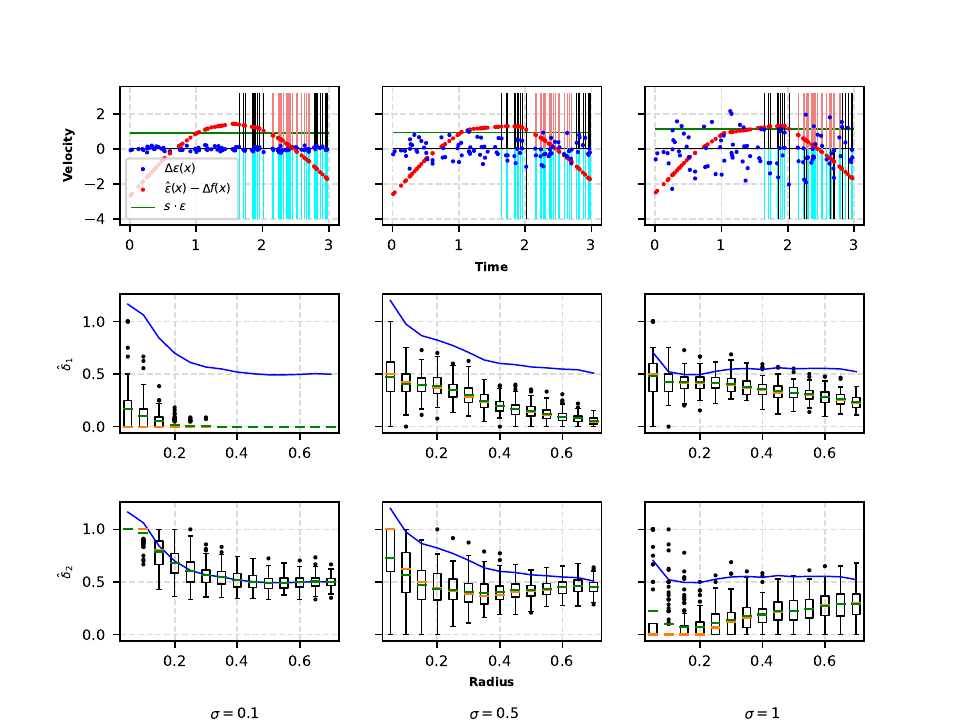}

	\caption{Using the same setup as in Fig.~\ref{fig:criteria_noise}, this shows the results for $X_n=2.3$. Here, the range of the radius is chosen as $r\in[1.6, 3.0]$ in steps of $0.05$.}
	\label{fig:bapc_criteria_2.3_eta_1}

\end{figure}

\begin{figure}[h]
	\centering
	\includegraphics[width=\textwidth]{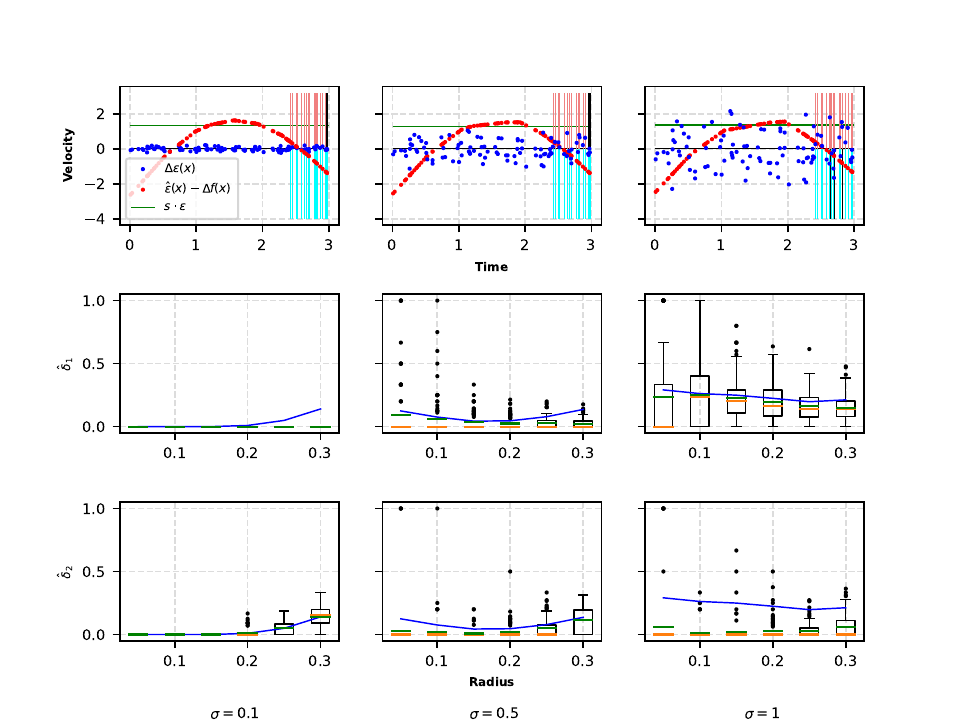}

	\caption{Using the same setup as in Fig.~\ref{fig:criteria_noise}, this shows the results for $X_n=2.7$. Here, the range of the radius is chosen as $r\in[2.4, 3.0]$ in steps of $0.05$.}
	\label{fig:bapc_criteria_2.7_eta_1}
\end{figure}

\begin{figure}[h]
	\centering
	\includegraphics[width=\textwidth]{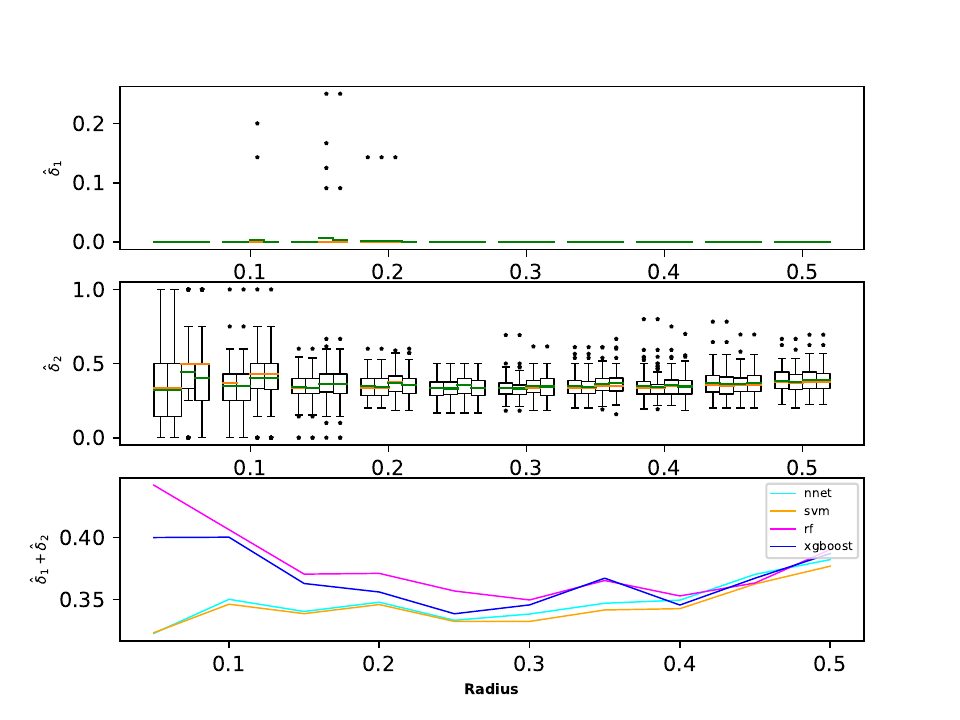}
	\caption{Using the same setup as in Fig.~\ref{fig:diffAI_0.5}, this show the results for $\sigma=0.1$.
	}
	
	\label{fig:diffAI_0.1}
\end{figure}

\begin{figure}[h]
	\centering
	\includegraphics[width=\textwidth]{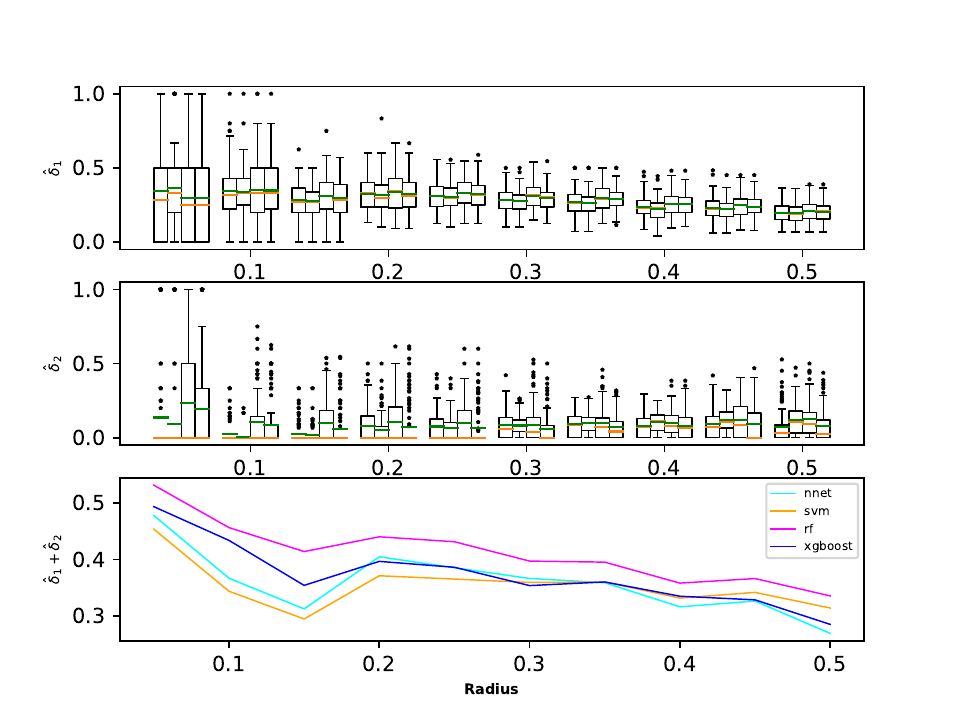}
	\caption{Using the same setup as in Fig.~\ref{fig:diffAI_0.5}, this show the results for $\sigma=1$.
	}
	
	\label{fig:diffAI_1.0}
\end{figure}

We repeated the simulated data from Sec.~\ref{sec:drag_descr} and the setup from Fig.~\ref{fig:criteria_noise} for $X_n=2.3$ and $X_n=2.7$. The results are shown in Fig.~\ref{fig:bapc_criteria_2.3_eta_1} and \ref{fig:bapc_criteria_2.7_eta_1}. We observe a similar behavior as for $X_n=2.5$, in particular $\hat\delta_1$ decreases while $\hat\delta_2$ increases with increasing $\U(X_n)$. The experiment with a comparison of different AI-models and $U=\N$ has been repeated for the noise levels $\sigma=0.1$ and $\sigma=1.0$ (see Fig.~\ref{fig:diffAI_0.1} and Fig.~\ref{fig:diffAI_1.0}).

\clearpage
\bibliographystyle{plain} 
\bibliography{references}

\begin{thebibliography}{10}

\bibitem{Arrieta}
Alejandro~Barredo Arrieta, Natalia D{\'\i}az-Rodr{\'\i}guez, Javier Del~Ser,
  Adrien Bennetot, Siham Tabik, Alberto Barbado, Salvador Garc{\'\i}a, Sergio
  Gil-L{\'o}pez, Daniel Molina, Richard Benjamins, et~al.
\newblock Explainable artificial intelligence (xai): Concepts, taxonomies,
  opportunities and challenges toward responsible ai.
\newblock {\em Information Fusion}, 58:82--115, 2020.

\bibitem{epidemiology}
J.~M. Bateson, T. F.;~Wright.
\newblock Regression calibration for classical exposure measurement error in
  environmental epidemiology studies using multiple local surrogate exposures.
\newblock {\em American Journal of Epidemiology}, 172:344--352, 2010.

\bibitem{failureControl}
A.~Ben~Abdessalem, A.; El-Hami.
\newblock A probabilistic approach for optimising hydroformed structures using
  local surrogate models to control failures.
\newblock {\em International Journal of Mechanical Sciences}, 96-97:143--162,
  2015.

\bibitem{Carvalho2019}
Diogo~V Carvalho, Eduardo~M Pereira, and Jaime~S Cardoso.
\newblock Machine learning interpretability: A survey on methods and metrics.
\newblock {\em Electronics}, 8(8):832, 2019.

\bibitem{cancer}
Abigail~S Caudle, Tse-Kuan Yu, Susan~L Tucker, Isabelle Bedrosian, Jennifer~K
  Litton, Ana~M Gonzalez-Angulo, Karen Hoffman, Funda Meric-Bernstam, Kelly~K
  Hunt, Thomas~A Buchholz, et~al.
\newblock Local-regional control according to surrogate markers of breast
  cancer subtypes and response to neoadjuvant chemotherapy in breast cancer
  patients undergoing breast conserving therapy.
\newblock {\em Breast Cancer Research}, 14:1--10, 2012.

\bibitem{productionOptimization}
Guodong Chen, Kai Zhang, Liming Zhang, Xiaoming Xue, Dezhuang Ji, Chuanjin Yao,
  Jun Yao, and Yongfei Yang.
\newblock Global and local surrogate-model-assisted differential evolution for
  waterflooding production optimization.
\newblock {\em SPE Journal}, 25(01):105--118, 2020.

\bibitem{Olivan}
Alberto Diez-Olivan, Javier Del~Ser, Diego Galar, and Basilio Sierra.
\newblock Data fusion and machine learning for industrial prognosis: Trends and
  perspectives towards industry 4.0.
\newblock {\em Information Fusion}, 50:92--111, 2019.

\bibitem{Dosilovic2018}
Filip~Karlo Do{\v{s}}ilovi{\'c}, Mario Br{\v{c}}i{\'c}, and Nikica Hlupi{\'c}.
\newblock Explainable artificial intelligence: A survey.
\newblock In {\em 2018 41st International convention on information and
  communication technology, electronics and microelectronics (MIPRO)}, pages
  0210--0215. IEEE, 2018.

\bibitem{Sahra2021}
Sahra Ghalebikesabi, Lucile Ter-Minassian, Karla DiazOrdaz, and Christopher~C.
  Holmes.
\newblock On locality of local explanation models.
\newblock In {\em Thirty-Fifth Conference on Neural Information Processing
  Systems}, volume~34, pages 18395--18407, 2021.

\bibitem{DARPA}
D~Gunning.
\newblock Explainable artificial intelligence (xai) darpa-baa-16-53.
\newblock {\em Defense Advanced Research Projects Agency}, 2016.

\bibitem{Haussler}
David Haussler.
\newblock Decision theoretic generalizations of the pac model for neural net
  and other\ learning applications.
\newblock {\em Information and Computation}, 100:78--150, 1992.

\bibitem{metrics}
Robert~R. Hoffman, Shane~T. Mueller, Gary Klein, and Jordan Litman.
\newblock Metrics for explainable ai: Challenges and prospects, 2019.

\bibitem{khan}
Maqbool Khan, Arshad Ahmad, Florian Sobieczky, Mario Pichler, Bernhard~A Moser,
  and Ivo Bukovsk{\`y}.
\newblock A systematic mapping study of predictive maintenance in smes.
\newblock {\em IEEE Access}, 10:88738--88749, 2022.

\bibitem{PdM}
Vikram Krishnamurthy, Kusha Nezafati, Erik Stayton, and Vikrant Singh.
\newblock Explainable ai framework for imaging-based predictive maintenance for
  automotive applications and beyond.
\newblock {\em Data-Enabled Discovery and Applications}, 4:1--15, 2020.

\bibitem{shapeDesign}
Leifur Leifsson, Elvar Hermannsson, and Slawomir Koziel.
\newblock Optimal shape design of multi-element trawl-doors using local
  surrogate models.
\newblock {\em Journal of Computational Science}, 10:55--62, 2015.

\bibitem{regression}
Simon Letzgus, Patrick Wagner, Jonas Lederer, Wojciech Samek, Klaus-Robert
  Müller, and Gregoire Montavon.
\newblock Toward {Explainable} {AI} for {Regression} {Models}.
\newblock {\em IEEE Signal Process. Mag.}, 39(4):40--58, July 2022.
\newblock arXiv:2112.11407 [cs, stat].

\bibitem{John2019}
Alex~John London.
\newblock Artificial intelligence and black-box medical decisions: accuracy
  versus explainability.
\newblock {\em Hastings Center Report}, 49(1):15--21, 2019.

\bibitem{alfredo}
Alfredo Lopez.
\newblock Explainable ai time series forecasting using a local surrogate model.
\newblock Presentation at the ENBIS 23 in Valencia, Spain, on 12 Sep 2023.

\bibitem{Bergman}
Klaus L\"uders and Gebhard von Oppen.
\newblock {\em Bergmann, Schaefer - Experimentalphysik 1. Mechanik - Akkustik -
  W\"arme}.
\newblock de Gruyter, 2008.

\bibitem{Molnar}
Christoph Molnar.
\newblock {\em Interpretable machine learning}.
\newblock leanpub.com, 2019.

\bibitem{Neugebauer2021}
Simon Neugebauer, Lukas Rippitsch, Florian Sobieczky, and Manuela Geiß.
\newblock Explainability of ai-predictions based on psychological profiling.
\newblock {\em Procedia Computer Science}, 180:1003--1012, 2021.
\newblock Proceedings of the 2nd International Conference on Industry 4.0 and
  Smart Manufacturing (ISM 2020).

\bibitem{Trust}
Marco Ribeiro, Sameer Singh, and Carlos Guestrin.
\newblock Why should {I} trust you?: Explaining the predictions of any
  classifier.
\newblock In {\em Proceedings of the 2016 Conference of the North {A}merican
  Chapter of the Association for Computational Linguistics: Demonstrations},
  pages 97--101, San Diego, California, 2016. Association for Computational
  Linguistics.

\bibitem{Ribeiro2016}
Marco~Tulio Ribeiro, Sameer Singh, and Carlos Guestrin.
\newblock Model-agnostic interpretability of machine learning.
\newblock 2016.

\bibitem{borehole}
M~Shahriari, D~Pardo, B~Moser, and F~Sobieczky.
\newblock A deep neural network as surrogate model for forward simulation of
  borehole resistivity measurements.
\newblock {\em Procedia Manufacturing}, 42:235--238, 2020.

\bibitem{Koiran}
Pascal Koiran; Eduardo~D. Sontag.
\newblock Vapnik-chervonenkis dimension of recurrent neural networks.
\newblock {\em Discrete Applied Mathematics}, 86:63--79, 1998.

\bibitem{Valiant}
L.~G. Valiant.
\newblock A theory of the learnable.
\newblock {\em Communications of the ACM}, 27:1134--1142, 1984.

\bibitem{Vapnik}
A.~Ya. Vapnik, V. N.;~Chervonenkis.
\newblock Necessary and sufficient conditions for the uniform convergence of
  means to their expectations.
\newblock {\em Theory of Probability \& Its Applications}, 26:532--553, 1982.

\end{thebibliography}





\end{document}